\documentclass{article}
\usepackage{spconf,amsmath,graphicx}
\usepackage{import}
\usepackage{multirow}
\usepackage{xcolor}

\date{}

\usepackage{enumitem}

\usepackage{url}

\usepackage{hyperref} 

\begin{document}

\onecolumn 

\begin{description}[labelindent=-1cm,leftmargin=1cm,style=multiline]

\item[\textbf{Citation}]{C. Zhou, G. AlRegib, A. Parchami, and K. Singh, "Learning Trajectory-Conditioned Relations to Predict Pedestrian Crossing Behavior," in \textit{IEEE International Conference on Image Processing (ICIP)}, Bordeaux, France, Oct. 16-19 2022.}


\item[\textbf{Review}]{Date of acceptance: June 20, 2022}

\item[\textbf{Codes}]{\url{https://github.com/olivesgatech/cross_predict}}



\item[\textbf{Copyright}]{\textcopyright 2022 IEEE. Personal use of this material is permitted. Permission from IEEE must be obtained for all other uses,
in any current or future media, including reprinting/republishing this material for advertising or promotional purposes,
creating new collective works, for resale or redistribution to servers or lists, or reuse of any copyrighted component of
this work in other works.}

\item[\textbf{Keywords}]{Intent prediction, Relation learning,
Pedestrian crossing} 

\item[\textbf{Contact}]{\href{mailto:chen.zhou@gatech.edu}{chen.zhou@gatech.edu} OR \href{mailto:alregib@gatech.edu}{alregib@gatech.edu}\\ \url{https://ghassanalregib.info/} \\
\url{http://alregib.ece.gatech.edu/} \\}
\end{description}

\thispagestyle{empty}
\newpage
\clearpage
\setcounter{page}{1}

\twocolumn


\title{Learning Trajectory-Conditioned Relations
to Predict Pedestrian Crossing Behavior}
%
\name{Chen Zhou$^{\star}$ \qquad Ghassan AlRegib$^{\star}$ \qquad Armin Parchami$^{\dagger}$ \qquad Kunjan Singh$^{\dagger}$}
\address{OLIVES at the Center for Signal and Information Processing\\School of Electrical and Computer Engineering\\
$^{\star}$Georgia Institute of Technology, Atlanta, GA 30332-0250\\
\{chen.zhou, alregib\}@gatech.edu\\
$^{\dagger}$Ford Motor Company, Dearborn, Michigan\\
\{mparcham, ksing114\}@ford.com}
%
%
%
%
\maketitle
\begin{abstract}
In smart transportation, intelligent systems avoid potential collisions by predicting the intent of traffic agents, especially pedestrians. Pedestrian intent, defined as future action, e.g., start crossing, can be dependent on traffic surroundings. In this paper, we develop a framework to incorporate such dependency given observed pedestrian trajectory and scene frames. Our framework first encodes regional joint information between a pedestrian and surroundings over time into feature-map vectors. The global relation representations are then extracted from pairwise feature-map vectors to estimate intent  with past trajectory condition. We evaluate our approach on two public datasets and compare against two state-of-the-art approaches. The experimental results demonstrate that our method helps to inform potential risks during crossing events with 0.04 improvement in F1-score on JAAD dataset and 0.01 improvement in recall on PIE dataset. Furthermore, we conduct ablation experiments to confirm the contribution of the relation extraction in our framework. 

\end{abstract}
\begin{keywords}
Intent Prediction, Relation Learning, Pedestrian Crossing 
\end{keywords}
\section{Introduction}
\label{sec:intro}

Intelligent systems are at a crossroad between perception and prediction. There has been significant advance on tackling the challenges of perception using learning-based algorithms. For instance, robust models \cite{prabhushankar2021extracting, lee2021open, kwon2020novelty, lehman2020structures} have been developed for image perception-based applications under challenging conditions.  In addition,  the success of  learning-based models on natural images has fostered its usage on computational image analysis including seismic \cite{benkert2021explaining, mustafa2021man} and medical \cite{temel2019relative} fields. While perception algorithms endows an intelligent system with the ability to see the current surroundings, prediction on future states informs intent and suggests safe maneuver.

\begin{figure}[th]
\begin{minipage}[b]{1.0\linewidth}
  \centering
  \centerline{\includegraphics[width=8.5cm]{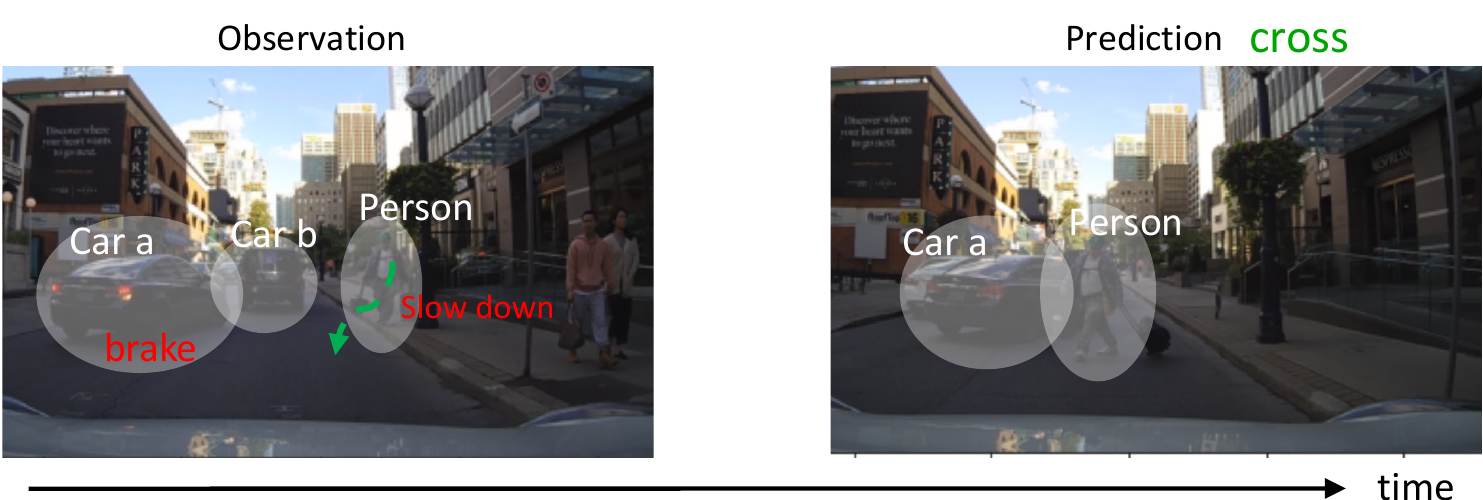}}
\end{minipage}
\caption{Example frames from the PIE dataset where a pedestrian intends to cross while interacting with traffic objects. The left figure shows a pedestrian 1) passes the vehicle b at further distance 2) 
slows down as the closer vehicle a brakes. The observed trajectory is shown in green dashed curve.}
\label{fig:intro-interact}
%
\end{figure}
Predicting future states of traffic agents, especially pedestrians, in complex urban scenes is safety-critical for intelligent transportation systems. The future states of pedestrians can be in the form of upcoming actions \cite{rasouli2020pedestrian}, e.g., road crossing. In this paper, we consider predicting future crossing as pedestrian intent prediction. 


Prediction can be challenging when pedestrian's future crossing depends on other traffic agents and environment. Traffic agents constantly avoid collisions by interacting within their local neighborhood as illustrated in Fig. \ref{fig:intro-interact}. Looking into local interaction at different regions, e.g., pedestrian $\leftrightarrow$ car a and pedestrian $\leftrightarrow$ car b together informs how their behavior change from one neighborhood to another across time from a global respective. Hence, we hypothesize that learning relation from spatiotemporal region-wise interaction with respect to a target agent informs intent. We develop a framework that learns relation from pedestrian visual interaction and motion dynamics to estimate the probability of binary crossing intent. The contributions of this paper include:
\begin{enumerate}
    \item We develop a relation extraction module that infers relation between pedestrian and surroundings from  region-wise local interaction.
    \item We show that integrating region-wise relation learning into prediction framework achieves competitive performance against existing methods. The improvement in F1-score and recall suggests lower ratio of missed crossing predictions.
\end{enumerate}


\section{Related Work}
\label{sec:format}
\textbf{Pedestrian Intent prediction} 
The objective of pedestrian intent prediction, specifically crossing prediction, is to anticipate whether a target pedestrian will cross the road at some time in the future. \cite{saleh2019real} extends DenseNet to extract spatio-temporal features from image sequences to estimate pedestrians’ crossing actions. Many algorithms rely on various input modalities and architectures \cite{rasouli2017they, rasouli2020pedestrian, rasouli2020multi}. For instance, the framework in \cite{rasouli2017they} utilizes traffic scene features and pedestrians looking and walking attributes by convolutional networks to forecast crossing action. Some recent methods utilize recurrent modeling \cite{rasouli2020pedestrian, bouhsain2020pedestrian}. The authors in \cite{rasouli2020pedestrian} develop a framework with different input modalities including pedestrian trajectories, poses, traffic scenes, and fuse them hierarchically within a stacked recurrent model. 

\textbf{Relation Modeling}
Some studies model the dependency between pedestrians and  traffic surroundings with object-level representations. For instance, the authors in \cite{yao2021coupling} design a module to extract object-wise interaction between various traffic objects based on their pre-detected location and visual states. Different from their method, we develop a relation extraction module that aggregates region-wise joint interaction to inform prediction.  


\section{Technical Approach}
\label{sec:pagestyle}
In this section, we introduce our framework to predict the probability of pedestrian crossing intent with region-wise relation representations and motion dynamics. We formalize the intent prediction problem in {subsection {3}.1}. We describe how relation representations with respect to motion dynamics are learned through our framework in {subsection {3}.2}.
\subsection{Problem Definition}
We define the pedestrian intent prediction as binary classification that predicts whether a target pedestrian will cross the road in the future based on the observation data. Without loss of generality, we assume that the observed trajectory of a target pedestrian is obtained from an object detection and tracking framework in advance. Let $X_i = [X_i^{t-\tau+1},..., X_i^{t}]$ denote the past trajectory of $i^{th}$ pedestrian from time $t-\tau+1$ to time $t$, where $X_i^{t}=(u_i^t, v_i^t, w_i^t, h_i^t)\in{\rm I\!R}^4 $ denotes the bounding box center, width, height of $i^{th}$ pedestrian at time $t$. Given the past trajectory $X_i$ along with the observed frames $I=[I^{t-\tau+1},..., I^{t}]$, we estimate the $i^{th}$ pedestrian's crossing probability $\hat{y}_i$.
\begin{figure}[th]
\begin{minipage}[b]{1.0\linewidth}
  \centering
  \centerline{\includegraphics[width=7.5cm]{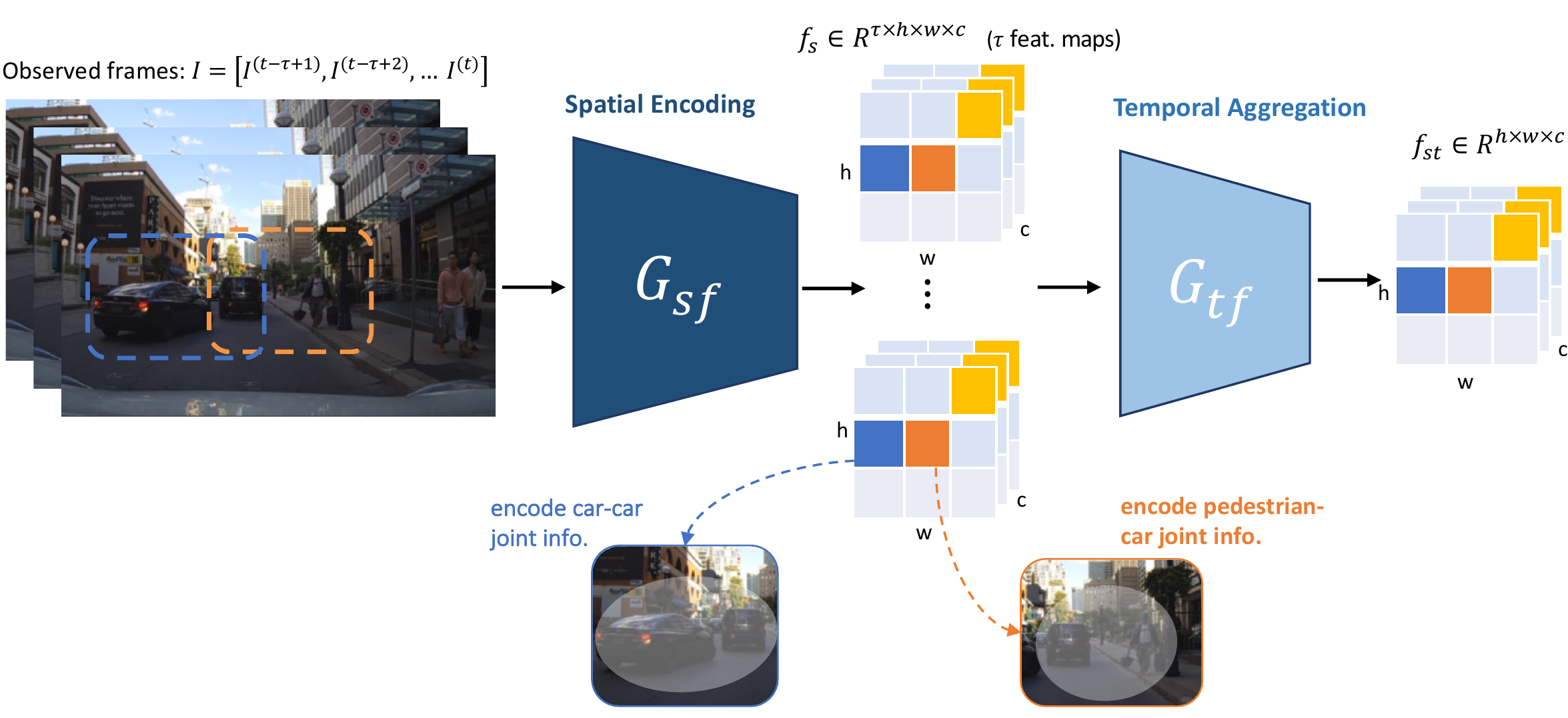}}
\end{minipage}
\caption{Joint information between object and surroundings is encoded within feature maps  by a visual encoder.}
\label{fig:ratp-visual-encoder}
%
\end{figure}
\subsection{Framework Overview}
Our prediction framework consists of three major components:  ({1}) a visual encoder to extract region-wise interaction evolving along time; ({2}) a relation extraction module that jointly relate local interaction from different regions to obtain relation from a global perspective; ({3}) a trajectory recurrent encoder that encodes pedestrian's observed motion dynamics. We elaborate each  component in more details in the following.

\textbf{Visual Encoder}
We employ a visual convolutional encoder that extracts local joint interaction between object and surroundings from observation frame sequence $I$ as shown in Fig.\ref{fig:ratp-visual-encoder}. The visual encoder can be divided into two parts:  ({1}) a {spatial module} that encodes frame-level joint local interaction within feature-map vectors; ({2})  a {temporal module} that aggregates frame-level feature-map vectors to extract spatiotemporal local interaction. 
The spatial module takes  $\tau$  observed frames $I=[I^{t-\tau+1},..., I^{t}]$  as inputs and passes them through a 2D ResNet34 \cite{he2016deep} backbone $G_{sf}$ to generate frame-level features
 $f_s\in {\rm I\!R}^{\tau\times h\times w\times c} $. The feature vectors (cells) in $f_s$ encode joint interaction within local regions, i.e, the feature vector in orange encodes pedestrian-car information. To capture temporal changes of joint information locally, we pass $f_s$ through a temporal module $G_{tf}$ to generate spatiotemporal local interaction representations $f_{st}$.
 Specifically, we employ $\tau\times1\times1$ convolutional 3D filters with $c=$512 channels to aggregate cells $f_s^{(k)}\in {\rm I\!R}^{\tau\times 1\times 1\times c}$ in  $f_s$ along temporal axis to generate $f_{st}\in {\rm I\!R}^{h\times w\times c}$, within which a cell is a $c$-dimensional spatiotemporal feature vector $f_{st}^{(k)}\in {\rm I\!R}^{c}$.

\begin{figure*}[t]
\begin{minipage}[b]{1.0\linewidth}
  \centering
  \centerline{\includegraphics[width=14.5cm]{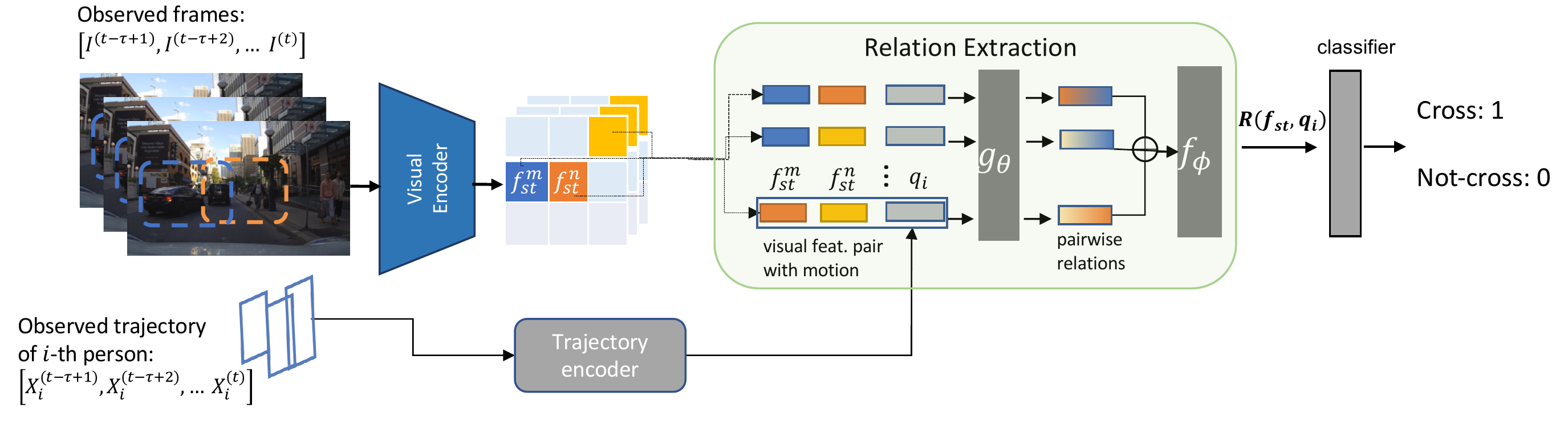}}
\end{minipage}
\caption{{The detailed architecture of our intent prediction framework. Each pairwise relation between $f_{st}^m$ and $f_{st}^n$ is conditioned on past agent-specific motion states $q_i$.}}
\label{fig:RATP-detailed-arch}
\end{figure*}
\textbf{Relation Extraction}
For a target pedestrian, relating local interaction from different cells $f_{st}^{(k)}$ (e.g., car-car and pedestrian-car) informs how behavioral interaction changes from one local region to another when objects moving around in the scene. While the visual encoder encodes spatiotemporal local interaction representations $f_{st}^{(k)}$, it less explicitly relate region-wise representations to jointly learn relation. Hence, we employ a relation extraction module to extract global-wise relation from
feature vectors $f_{st}^{(1)},...f_{st}^{(k)},...,f_{st}^{(h\times w)}$ in $f_{st}$. Inspired by \cite{santoro2017simple}, we extend the notion of entity to a spatiotemporal feature representation encoded from a local region across time, which is  $f_{st}^{(k)}$ in our case.
Hence, we define an entity set as $f_{st} = \{f_{st}^{(1)},...f_{st}^{(k)},...,f_{st}^{(h\times w)}\}$. 
Following such definition,  we learn the relation between entities in $f_{st}$ via a composite function $R(*)$ as following:
\begin{equation}
    R(f_{st}) = f_\phi \left(\sum_{m,n}g_\theta \left( f_{st}^m, f_{st}^n \right)\right)
\end{equation}
where $g_\theta(*)$ denotes a pair-wise relation function that takes a pair of feature vectors $f_{st}^m, f_{st}^n$ in  $f_{st}$ as input to learn their relation. The outcomes of $g_\theta(*)$ from all feature vector pairs are then aggregated to be passed through  $f_\phi(*)$ to infer relations globally. In general, $f_\phi(*)$ and $g_\theta(*)$ can be simple neural network functions, e.g., multi-layer perceptrons (MLPs). We use one linear layer as $g_\theta$ and an identity function as $f_\phi$. In the high level, $R(*)$ receives pair-wise features that represents region-wise local interaction, if any, and integrate them to form global relation. 

We introduced how to model global relation from region-wise interactions via a relation module. We then elaborate on how we incorporate extracted relation representations $R(f_{st})$ into our prediction framework.

\textbf{Trajectory Encoding} The past observation of pedestrian's bounding box implicitly captures the motion dynamics in relative distance to the surroundings, which is informative to relation learning. Thus, We employ a recurrent encoder to extract motion dynamics of a target pedestrian. The encoder consists of gated recurrent union (GRU) with hidden size 256 that generates motion hidden  states $q_i$ from i-th pedestrian's trajectory $X_i$, as shown in the lower part of Fig. \ref{fig:RATP-detailed-arch}. 


The relation features $R(f_{st})$ are agent-agnostic representations. Given the same observation frames $I$ with multiple pedestrians and traffic objects, the intentions of different pedestrians can be influenced by different relation context. To learn agent-specific relation context  for each target pedestrian,  $R(f_{st})$ should be conditioned on pedestrian's historical motion dynamics. Hence, we condition relation learning on $q_i$ for $i$-th pedestrian and use the trajectory-conditioned relation features $R(f_{st}, q_i)$ to predict intent, as shown in the relation extraction block in {Fig.~\ref{fig:RATP-detailed-arch}}.
Specifically, we concatenate $q_i$ to every concatenated pair of  $f_{st}^m, f_{st}^n$ and pass triplets $(f_{st}^m, f_{st}^n, q_i)$ to $g_\theta$ and $f_\phi$ to obtain $R(f_{st}, q_i)$. We append a MLP as classifier to estimate the intent probability $\hat{y}_i$ given relation features $R(f_{st}, q_i)$.

\textbf{Learning Objectives} We employ the binary cross-entropy objective for intent estimation $\hat{y}_i$ and ground-truth intent $y_i$ for each pedestrian. The loss function can be expressed as following:
\begin{equation}
		\mathcal{L} = - \frac{1}{N} \sum_{i=1}^{N} (y_i\log{\hat{y}_i} + (1-y_i)\log({1-\hat{y}_i}))
\end{equation}
where $N$ denotes the total number of training pedestrians.


\section{Experiments}
\label{sec:typestyle}
In this section, we empirically evaluate our intent prediction framework on two public datasets: Joint Attention for Autonomous Driving (JAAD) \cite{kotseruba2016joint} and Pedestrian Intention Estimation (PIE) \cite{rasouli2019pie} that contain scenarios where pedestrians approaching the road and might cross in the future. We provide a comparative study to demonstrate the efficacy of our approach and an ablation discussion on relation extraction.
\def\arraystretch{1}
\begin{table*}[th]
\caption{Results comparison  of our approach with other methods on JAAD and PIE datasets.}
\begin{center}
\begin{tabular}{|c|ccccc|ccccc|}
\hline
\multirow{2}{*}{Methods} & \multicolumn{5}{c|}{PIE}                   & \multicolumn{5}{c|}{JAAD}            \\ \cline{2-11}
  & Acc & AUC &F1 &P & R     & Acc & AUC &F1 &P & R
\\ \hline 
 
 SF-GRU\cite{rasouli2020pedestrian}     & 0.82  & 0.79 & 0.69 & 0.67 & 0.70   & 0.84 & 0.84 & 0.65 & 0.54 & \textbf{0.84}\\ \cline{1-11} 
  ARN\cite{yao2021coupling}     & \textbf{0.84} & \textbf{0.88} & \textbf{0.90} & \textbf{0.96} & 0.85   & \textbf{0.87} & \textbf{0.92} & 0.70 & 0.66 & 0.75\\ \cline{1-11} 
  Ours (w/o relation)     & 0.82  & 0.81  & 0.84 & 0.89 & 0.80   & 0.81 & 0.84 & 0.70 & 0.63 & 0.82\\ \cline{1-11} 
                          Ours   & \textbf{0.84}  & 0.85 & \textbf{0.90} & 0.94 & \textbf{0.86}  & 0.84  & 0.89 & \textbf{0.74} & \textbf{0.67} & \textbf{0.84}\\ \hline
\end{tabular}
\end{center}
\label{Tab:T1}
\end{table*}
\subsection{Datasets and Experimental Setting}
\textbf{Datasets} 
The two datasets contain video sequences collected in urban driving scenes. JAAD contains 323 video clips with total 2786 pedestrians. PIE contains long videos from 6 sets covering 6 hours of driving footage with total 1842 pedestrians. There are 2100 not-crossing pedestrians and 686 crossing pedestrians in the JAAD dataset. PIE  contains 1322 non-crossing pedestrians and 512 crossing ones. The pedestrian tracks in both datasets are annotated at 30Hz.

\begin{figure}[thb]
\begin{minipage}[b]{1.0\linewidth}
  \centering
  \centerline{\includegraphics[width=4.5cm]{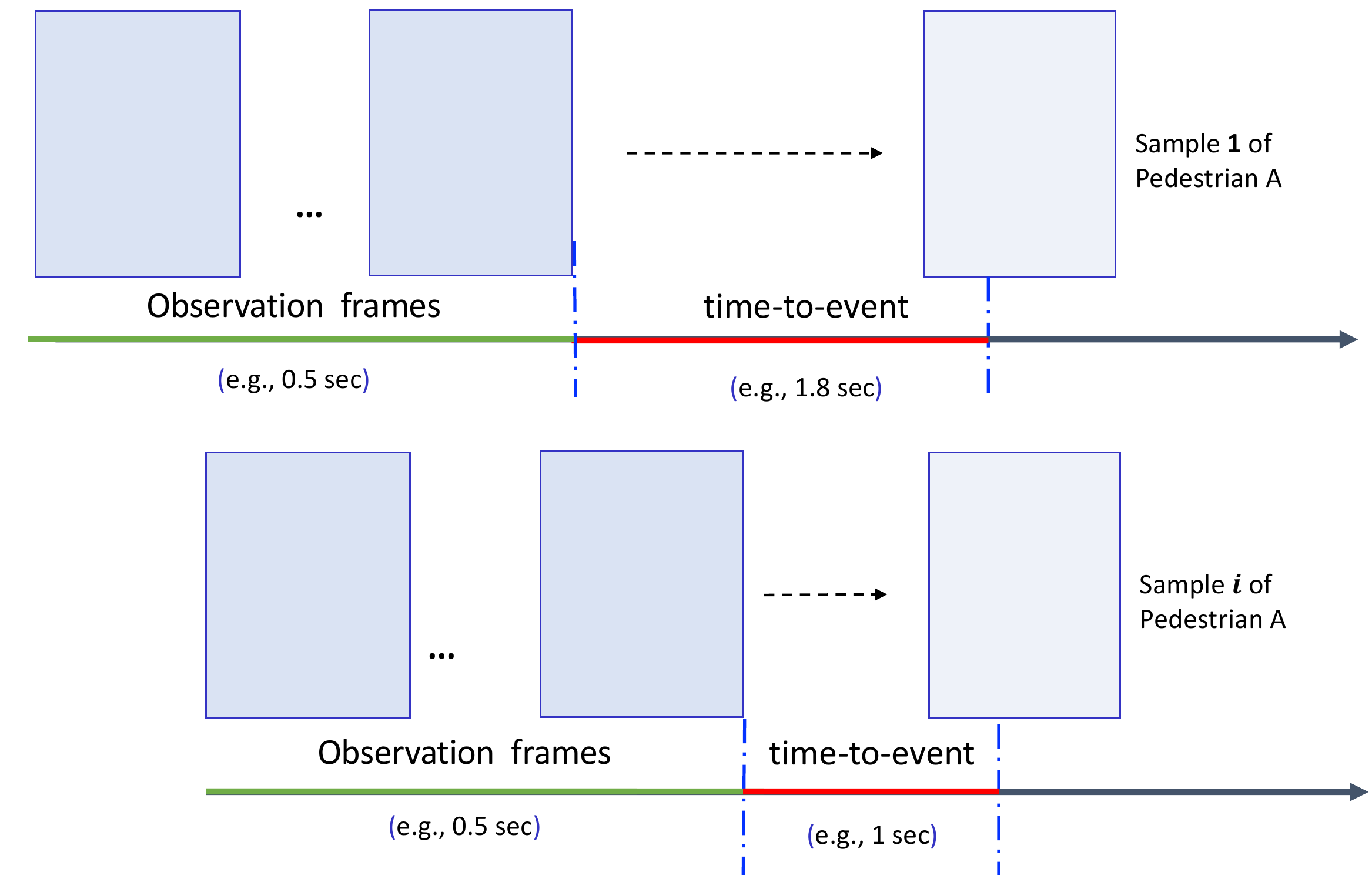}}

\end{minipage}
\caption{Subsampling for pedestrian crossing prediction: sample observation frames prior to the starting point of crossing. We subsample the observation data by varying time-to-event.}
\label{fig:Setting-subsample}
\end{figure}

\textbf{Experimental Setting}
Throughout our experiments, we follow the same data preprocessing procedure and evaluation protocol as previous work \cite{rasouli2019pie, rasouli2020pedestrian}. We use the standard data splits of JAAD as in \cite{rasouli2018s} where the dataset is divided into 177 training clips, 29 validation clips and 117 test clips, respectively. The pedestrian tracks are divided into 1355, 202 and 1023 for train/val/test splits in the JAAD data. For the PIE dataset, we follow the same data splits defined in \cite{rasouli2019pie}: videos from set01, set02 and set06 are used for training, set04 and set05 for validation and set03 for test. The number of pedestrian tracks in PIE is 880, 243 and 719 in training, validation and test sets.
To predict the crossing probability of a pedestrian, we subsample each pedestrian track into multiple observation sequences with fixed length of $\tau=$16 frames (0.5s observation). The sample overlap ratio is set to 0.6 for PIE and 0.8 for JAAD dataset. The last frame of each subsampled observation sequence falls between 30 and 60 frames (or 1s -- 2s) prior to the annotated starting point of a crossing event, as illustrated in Fig. \ref{fig:Setting-subsample}. Such time-to-event (TTE) setting allows prediction beforehand as some time should be allotted for emergency maneuver. 

\textbf{Evaluation Metrics.}
To evaluate the performance of our prediction framework, we use several binary classification metrics: accuracy, area under the curve (AUC), precision, recall and F1-score.
\subsection{Results}
We compare our method with two other state-of-the-art prediction methods as following: 
\begin{enumerate}
    \setlength{\itemsep}{0pt}
    \setlength{\parskip}{0pt}
    \setlength{\parsep}{0pt}
    \item \textit{SF-GRU}\cite{rasouli2020pedestrian}: a framework that hierarchically fuses features from trajectories, poses and traffic scenes  within a stacked recurrent model.
    \item \textit{ARN}\cite{yao2021coupling}: a framework that extracts object-wise interaction from object location and visual states.
\end{enumerate}

The evaluation results of state-of-the-art methods and ours are shown in Table \ref{Tab:T1}.  Our method outperforms SF-GRU model in several metrics, and achieves similar or superior performance in recall and F1-score on both datasets. Specifically, our model results in 0.04 increase in F1-score on JAAD dataset and 0.01 increase in recall on PIE dataset.
The improvement in recall and F1-score implies lower ratio of missed crossing predictions. Similar trend can be observed when comparing the fusion model SF-GRU and the interaction model ARN. This indicates that incorporating relation learning into prediction framework helps to alarm the intelligent system about potential risks. Our method yields lower acc and AUC on JAAD. The pedestrians in JAAD are close to the curbside with similar trajectories while less interaction with traffic infrastructures. The relation representations $R(f_{st}, q_i)$ on JAAD is majorly influenced by the less discriminative $q_i$. Hence, there are fewer True Negatives, leading to the drop of acc and AUC. 

To investigate the effect of relation extraction within our framework, we conduct an ablation experiment by removing the  relation extraction. We employ a model that consists of the same visual encoder, the same trajectory encoder and the same final classifier on the same inputs. However, we replace the relation extraction ($g_\theta$ and $f_\phi$) with an MLP with same number of layers. This MLP connects to the concatenated features of full feature maps $f_{st}$ and $q_i$, instead of triplets of vectors $(f_{st}^m, f_{st}^n, q_i)$. Thus, there are more parameters in this model.  We include the results of this non-relation model in the third row of the Table\ref{Tab:T1} as Ours (w/o relation). The improved performance by using relation extraction confirms its contribution in the estimation of pedestrian crossing intent. 

\section{Conclusion}
In this work, we develop a framework with trajectory-conditioned relation learning to predict pedestrian crossing behavior. A relation extraction module is developed to learn relations between a target pedestrian and surroundings from region-wise interaction through our framework. We evaluate our method on two public intent prediction datasets and compare against state-of-the-art approaches. The experimental results demonstrate that our framework helps to inform potential risks during crossing events  with 0.04 increase in F1-score on JAAD dataset and 0.01 increase in recall on PIE dataset. We further conduct ablation experiments to verify the contribution of the relation extraction in our framework.



\bibliographystyle{IEEEbib}
\bibliography{strings,refs}

\end{document}